\documentclass{article}
\usepackage{spconf,amsmath,graphicx,hyperref}
\usepackage{algorithmic}
\usepackage{algorithm}
\usepackage{array}
\usepackage{bbding}
\usepackage{textcomp}
\usepackage{stfloats}
\usepackage{url}
\usepackage{verbatim}
\usepackage{graphicx}
\usepackage{cite}
\usepackage[utf8]{inputenc} 
\usepackage[T1]{fontenc}    
\usepackage{booktabs}       
\usepackage{amsfonts}       
\usepackage{nicefrac}       
\usepackage{microtype}      
\usepackage{xcolor}         
\usepackage{wrapfig}
\usepackage{multirow}
\usepackage[table]{xcolor}
\usepackage{caption}
\usepackage{subcaption}
\usepackage{adjustbox}
\usepackage{colortbl} 
\arrayrulecolor{black}
\usepackage{xr}
\usepackage{mathtools}
\usepackage{amsthm}
\usepackage{siunitx}

\newcommand\tblue[1]{\textcolor[RGB]{0,0,180}{#1}}
\newcommand\tgreen[1]{\textcolor[RGB]{0,130,0}{#1}}
\newcommand\tred[1]{\textcolor[RGB]{150,0,0}{#1}}
\title{PRISM: Precision-Recall Informed Data-Free Knowledge \\
Distillation via Generative Diffusion}
\name{Xuewan He, Jielei Wang, Zihan Cheng, Yuchen Su, Shiyue Huang, Guoming Lu\textsuperscript{$\dagger$}
\thanks{\textsuperscript{$\dagger$} Corresponding author.}
}

\address{University of Electronic Science and Technology of China, Chengdu, China\\
}

\begin{document}
\ninept
\captionsetup[figure]{skip=2pt}
\captionsetup[table]{skip=2pt}
\maketitle
\begin{abstract}
Data-free knowledge distillation (DFKD) transfers knowledge from a teacher to a student without access to the real in-distribution (ID) data.
While existing methods perform well on small-scale images, they suffer from mode collapse when synthesizing large-scale images, resulting in limited knowledge transfer.
Recently, leveraging advanced generative models to synthesize photorealistic images has emerged as a promising alternative.
Nevertheless, directly using off-the-shelf diffusion to generate datasets faces the precision-recall challenges: 1) ensuring synthetic data aligns with the real distribution, and 2) ensuring coverage of the real ID manifold. 
In response, we propose PRISM, a precision–recall informed synthesis method. 
Specifically, we introduce Energy-guided Distribution Alignment to avoid the generation of out-of-distribution samples,
and design the Diversified Prompt Engineering to enhance coverage of the real ID manifold. 
Extensive experiments on various large-scale image datasets demonstrate the superiority of PRISM. 
Moreover, we demonstrate that models trained with PRISM exhibit strong domain generalization. 
\end{abstract}
%
\section{Introduction}
\label{sec:intro}
Knowledge distillation (KD)~\cite{hintonDistillingKnowledgeNeural2015} has emerged as a powerful paradigm for model compression, wherein the expertise of a pretrained teacher network is transferred to a more compact student network. 
KD approaches typically presuppose access to the original training data, which serve as indispensable guidance for effective student learning. 
In real-world scenarios, however, such data are often inaccessible or restricted due to privacy concerns, security protocols, or proprietary interests~\cite{bae2018security}.
To address this challenge, data-free knowledge distillation (DFKD) has emerged as an active area of research~\cite{chenDataFreeLearningStudent2019, yinDreamingDistillDataFree2020, fangContrastiveModelInversion2021, fang100xFasterDataFree2022, tranNAYERNoisyLayer2024}, leveraging model inversion to facilitate knowledge transfer from the teacher without dependence on the original training data.

\begin{figure}[t]
    \centering
    \centering
    \includegraphics[width=\linewidth]{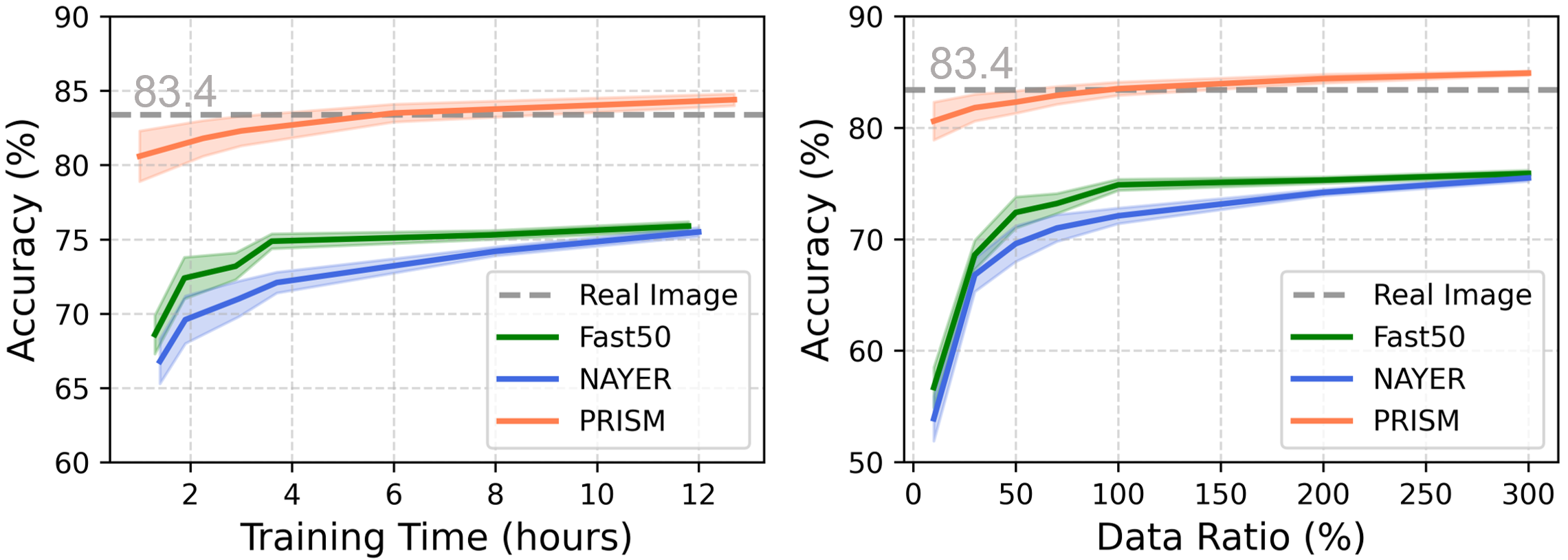}
    \caption{Comparison of accuracy versus training time (left) and data ratio (right) for our PRISM and Inversion-based methods.}
    \label{fig:efficiency}
    \vspace{-13pt}
\end{figure}

\begin{figure}[t]
    \centering
    \includegraphics[width=\linewidth]{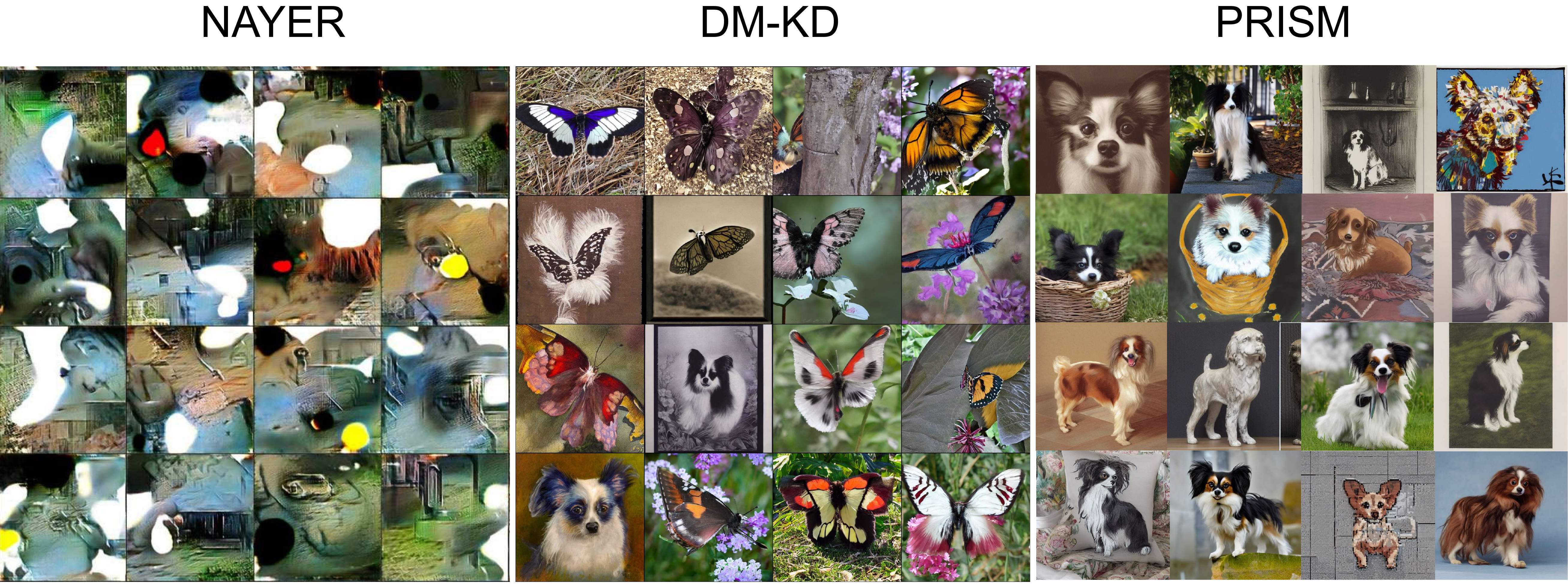}
    \caption{Visualization of the \textit{``papillon''} class on each method. \textit{``Papillon''} is a small, slender toy spaniel.
    The images synthesized by NAYER are homogeneous and have low information density; DM-KD misidentifies \textit{``Papillon''} as \textit{``Butterfly''}. 
    While our method is semantically correct and diverse.
    }
    \label{fig:visualize}
    \vspace{-10pt}
\end{figure}

Existing data-free knowledge distillation (DFKD) approaches~\cite{fangContrastiveModelInversion2021, fang100xFasterDataFree2022, tranNAYERNoisyLayer2024} predominantly rely on the batch normalization (BN) statistics~\cite{yinDreamingDistillDataFree2020} embedded in the teacher network to reconstruct surrogate data. 
While this paradigm demonstrates strong performance on small-scale benchmarks, it degrades notably when scaled to large and more realistic datasets.
The fundamental difficulty arises from the ill-posed nature of the inverse mapping from a compact output space to a high-dimensional image space~\cite{liMixMixAllYou2022}. 
For instance, on ImageNet, the task involves mapping a 1000-dimensional output of logits to a 224×224×3 image. Without real image priors, this mapping often results in images with significant noise, exhibiting a lack of semantic coherence and low information density, with examples shown in Fig.~\ref{fig:visualize}.
Moreover, adversarial-based~\cite{fangContrastiveModelInversion2021, fang100xFasterDataFree2022, tranNAYERNoisyLayer2024} variants are often plagued by mode collapse, where synthesis overly conforms to the teacher’s biases and produces highly homogeneous data. 
These limitations account for the pronounced performance drop of existing methods in large-scale, real-world scenarios, as illustrated in Fig.~\ref{fig:efficiency}.

\begin{figure*}[t]
    \centering
    \includegraphics[width=0.96\linewidth]{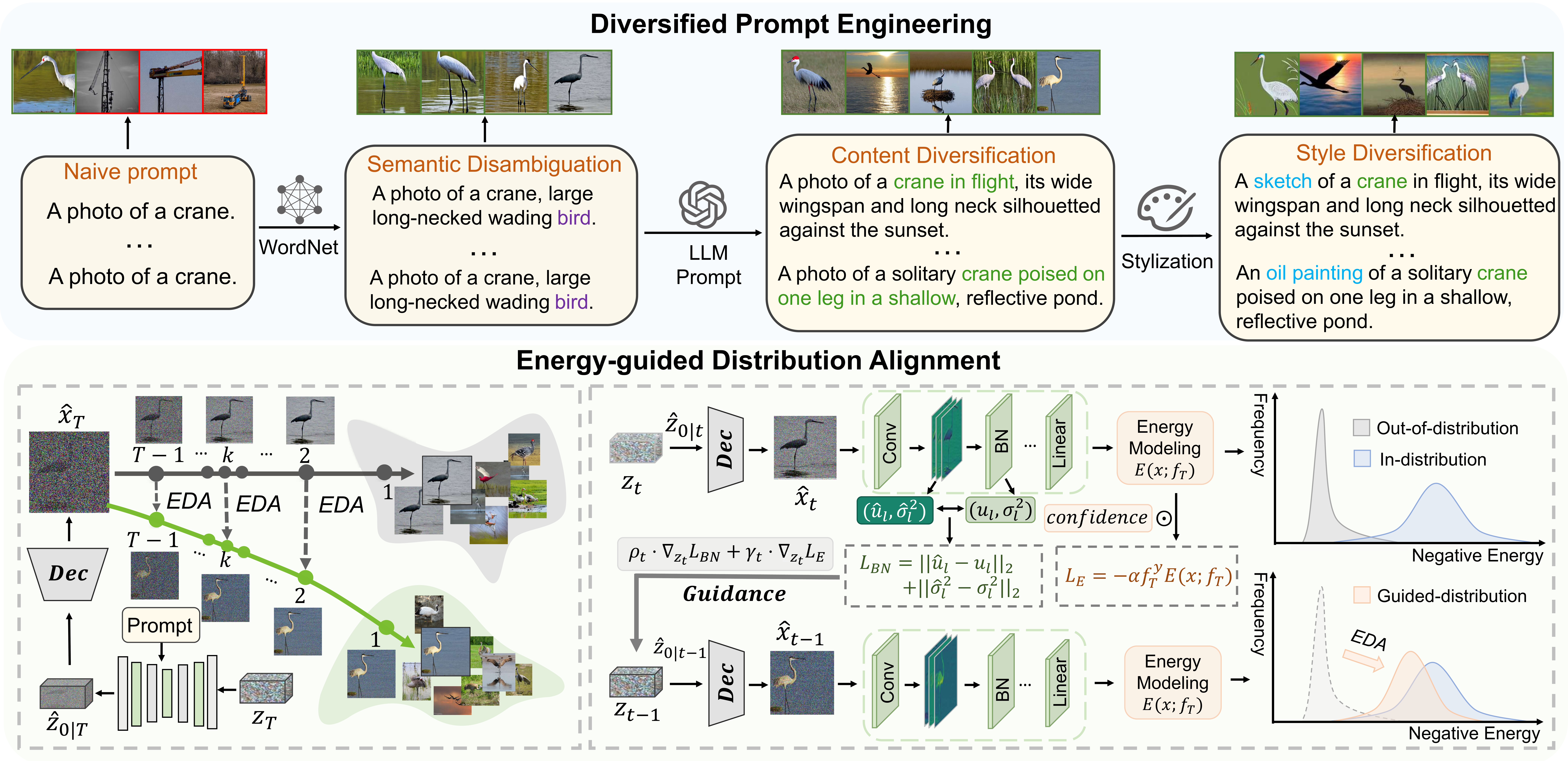}
    \caption{Overall framework of PRISM.
    The goal of PRISM is to maximize sample diversity while ensuring that the synthetic samples approximate the real ID manifold.
    PRISM involves first obtaining diverse cues through DPE, and then guiding the distribution through EDA in the diffusion reverse process, so that diverse synthetic samples fall within the real training distribution.
    } 
    \label{fig:method}
    \vspace{-10pt}
\end{figure*}

To overcome the adversity of large-scale image synthesis, recent studies~\cite{liSyntheticDataDiffusion2023,li2024towards,liGenQQuantizationLow2024} have leveraged diffusion models as powerful generative priors, thereby significantly enhancing the fidelity and efficiency of reconstructed data. 
However, directly leveraging diffusion models still faces fundamental precision–recall challenges, as the synthesized data often exhibits simultaneously low precision and low recall.
Specifically, due to the absence of real in-distribution (ID) data, the synthesized distribution of diffusion models often suffers from distributional shift, producing out-of-distribution (OOD) samples that result in low precision.
Moreover, when using text-to-image (T2I) diffusion models~\cite{saharia2022photorealistic}, previous methods only used the class priors to construct the prompt (e.g., “\textit{A photo of a <class>}”), which not only introduces semantic ambiguity (Fig.~\ref{fig:visualize} and \ref{fig:method}) but also restricts sample diversity. 
As a result, the reconstructed distribution fails to adequately cover the training data manifold, leading to insufficient recall.
Therefore, resolving the fundamental precision-recall challenge is essential to fully unlock the potential of diffusion models and enhance their contribution to data-free knowledge distillation.

In this paper, we propose PRISM, a \underline{P}recision–\underline{R}ecall \underline{I}nformed \underline{S}ynthesis \underline{M}ethod (Fig.~\ref{fig:method}), which consists of two complementary modules: Energy-guided Distribution Alignment (EDA) and Diversified Prompt Engineering (DPE), anchoring the improvement of Precision and Recall respectively.
EDA leverages the principle of diffusion model classifier guidance and incorporates the teacher’s prior knowledge to construct an energy function, where high-energy samples are treated as OOD data and low-energy samples as ID data. 
By minimizing this energy during the reverse process, the latent representations are guided toward the ID manifold, effectively enhancing the precision of synthesis.
Complementarily, BN statistics from the teacher are integrated to further strengthen distributional alignment, ensuring feature-level congruity.
Once EDA ensures distribution consistency, we employ DPE driven by large language models (LLMs) to improve recall.
Specifically, instead of directly using naive text prompts, DPE effectively enhances both semantic consistency and data diversity through semantic disambiguation, content diversification, and style diversification, thereby achieving a more comprehensive coverage of the original distribution manifold.
Together, our PRISM method synergistically improves both precision and recall through EDA and DPE, achieving an optimal trade-off in synthetic data distribution.

Extensive experiments across various ImageNet~\cite{russakovskyImageNetLargeScale2015} subsets demonstrate the effectiveness of our approach. 
As illustrated in Fig.~\ref{fig:efficiency}, PRISM exhibits superior data efficiency compared to inversion-based methods. 
Remarkably, with only 10\% synthetic data, PRISM achieves 80\% accuracy, approaching the 83.4\% attained on full data.
Additionally, we show in $\S$~\ref{sec:dg} that PRISM-trained students exhibit strong domain generalization, comparable to and even surpassing that of models trained from scratch on the training dataset.

%

Our major contributions are summarized as follows:
\begin{itemize}
    \item We analyzed the precision-recall challenges inherent in directly using diffusion models for synthesizing DFKD data, and proposed PRISM to bridge the gap between the synthesized data and the real data distribution.
    \item By integrating Energy-guided Distribution Alignment and Diversified Prompt Engineering, PRISM ensures alignment with the real ID distribution (precision) and promotes more comprehensive coverage (recall).
    \item Extensive evaluations confirm the superiority of PRISM, delivering state-of-the-art results across both inversion-based and diffusion-based DFKD baselines.
\end{itemize}
\vspace{-5pt}
\section{Methodology}
\subsection{Revisit to Diffusion-based Data-Free Knowledge Distillation}
Data-Free Knowledge Distillation (DFKD) aims to transfer knowledge from a teacher $f_{T}$ to a student $f_{S}$ without relying on the original dataset $\mathcal{X} \sim \mathcal{P}$. 
Formally, the objective is to minimize the KL divergence between teacher and student:
\begin{equation}
\mathbb{E}_{x \sim \mathcal{Q}(x)} \big[ \mathrm{KL}(f_T(x) \,\|\, f_S(x)) \big],
\quad \text{s.t.} \quad \mathrm{KL}(\mathcal{Q} \,\|\, \mathcal{P}) \leq \epsilon,
\end{equation}
where the synthetic distribution $\mathcal{Q}(x)$ is constrained to approximate the real distribution $\mathcal{P}(x)$. 
To achieve this, recent Diffusion-based DFKD methods \cite{liSyntheticDataDiffusion2023,li2024towards,liGenQQuantizationLow2024} synthesize surrogate data by using the powerful distribution priors of a pretrained text-to-image (T2I) latent diffusion model (LDM).
Formally, given a class name as text condition $c$, the reverse process of the LDM $\boldsymbol{\epsilon}_\phi$ first predicts the clean data point $\hat{\boldsymbol{z}}_{0|t}$ based on $\boldsymbol{z}_{t}$ as:
\vspace{-3pt}
\begin{equation}
\hat{\boldsymbol{z}}_{0|t} = \frac{1}{\sqrt{\alpha_t}} \left( \boldsymbol{z}_{t} - \sqrt{1-\alpha_t} \, \boldsymbol{\epsilon}_\phi(\boldsymbol{z}_{t}, t, c) \right).
\vspace{-3pt}
\end{equation}
$\boldsymbol{z}_{t-1}$ is then sampled from: 
\begin{algorithm}[h]
\caption{PRISM: A Precision–Recall Informed Synthesis Method}
\label{alg:prism}
\begin{algorithmic}[1]
\REQUIRE Pretrained teacher model $f_T$, LDM $\boldsymbol{\epsilon}_\phi$, VAE decoder $\mathcal{D}$, class set $\mathcal{Y}$, LLM $\mathcal{M}$
\ENSURE Synthetic dataset $\mathcal{X}_{\text{syn}}$
\STATE Generate diversified prompts $\mathcal{P}_{\text{all}}$ for all classes via DPE
\STATE $\mathcal{X}_{\text{syn}} \gets \emptyset$

\FOR{each prompt $(c,y) \in \mathcal{P}_{\text{all}}$}
    \STATE $\boldsymbol{z}_T \sim \mathcal{N}(0, I)$
    \FOR{$t = T$ \TO $1$}
        \STATE $\hat{\boldsymbol{z}}_{0|t} = \frac{1}{\sqrt{\alpha_t}} \left( \boldsymbol{z}_{t} - \sqrt{1-\alpha_t} \, \boldsymbol{\epsilon}_\phi(\boldsymbol{z}_{t}, t, c) \right)$
        \STATE $\boldsymbol{\hat{x}}_t = \mathcal{D}(\hat{\boldsymbol{z}}_{0|t})$
        \IF{$t \in [\tau_{\text{start}}, \tau_{\text{end}}]$}
            \STATE $\boldsymbol{z}_{t-1} = s(\boldsymbol{z}_t, t, \boldsymbol{\epsilon}_{\phi}) - \rho_t \nabla_{\boldsymbol{z}_t} \mathcal{L}_{\mathrm{BN}} - \gamma_t \nabla_{\boldsymbol{z}_t} \mathcal{L}_{\mathrm{E}}$
        \ELSE
            \STATE $\boldsymbol{z}_{t-1} = s(\boldsymbol{z}_t, t, \boldsymbol{\epsilon}_{\phi})$
        \ENDIF
    \ENDFOR
    \STATE $\boldsymbol{x}_{\text{syn}} = \mathcal{D}(\boldsymbol{z}_0)$
    \STATE $\mathcal{X}_{\text{syn}} \gets \mathcal{X}_{\text{syn}} \cup \{(\boldsymbol{x}_{\text{syn}}, y)\}$
\ENDFOR
\end{algorithmic}
\end{algorithm}
\vspace{-3pt}
\begin{equation}
\mathcal{N}\!\left(\sqrt{\alpha_{t-1}} \hat{\boldsymbol{z}}_{0|t} + \sqrt{1-\alpha_{t-1}-\sigma_t^2}\,\boldsymbol{\epsilon}_\phi(\boldsymbol{z}_{t}, t, c), \; \sigma_t^2 I \right),
\vspace{-3pt}
\end{equation}
where $\sigma_t$ is the predefined noise factor. For notation simplicity, we abstract this process as: 
$\boldsymbol{z}_{t-1} = s(\boldsymbol{z}_{t}, t, \boldsymbol{\epsilon}_\phi)$.
However, this process does not leverage any prior knowledge in the original training data, which might cause the synthetic data to be OOD relative to the training data.
Moreover, using a naive text prompt (e.g., \textit{“A photo of a <class>”}) restricts diversity and risks semantic ambiguity, as polysemous class names (e.g., \textit{“crane”}) can correspond to entirely different visual concepts, such as a bird or a machine.
\subsection{Energy-guided Distribution Alignment}
To guide the latent $\boldsymbol{z}_{t}$ so that the synthesized data can fall within the real ID manifold, we further introduce another conditional guidance in addition to the text guidance. 
We can then rewrite the reverse process to obtain $\boldsymbol{z}_{t-1}$ that meets the specified conditional constraints~\cite{ye2024tfg}:
\begin{equation}
    \boldsymbol{z}_{t-1} = s(\boldsymbol{z}_t, t, \boldsymbol{\epsilon}_{\phi})- \rho_t \nabla_{\boldsymbol{z}_t} f_C(\hat{\boldsymbol{z}}_{0|t}),
\end{equation}
where $f_C(\cdot)$ is the metric function of the predicted sample $\hat{\boldsymbol{z}}_{0|t}$ to the specific condition C.

To achieve ID data synthesis, we propose Energy-guided Distribution Alignment (EDA), which formulates this conditional guidance as an energy-based optimization objective.
Specifically, we introduce energy-based modeling~\cite{liu2020energy}, which models the synthetic distribution via the energy function $\mathbf{E}(\boldsymbol{\hat{x}}_t,f_T)$ as follows:
\vspace{-5pt}
\begin{equation}
\label{eq:energy}
\mathbf{E}(\boldsymbol{\hat{x}}_t,f_T)  = -\alpha \log \sum_{i}^{K} e^{f^i_T(\boldsymbol{\hat{x}}_t)/\alpha},
\vspace{-5pt}
\end{equation}
where $\boldsymbol{\hat{x}}_t = \mathcal{D}(\hat{\boldsymbol{z}}_{0|t})$, $\mathcal{D}$ is the VAE Decoder. 
The principle of EDA is that the pretrained teacher induces systematically lower energy values for ID data relative to OOD counterparts.
Accordingly, EDA guides diffusion to generate energy-minimized synthetic samples through the teacher model’s prior knowledge of the training data:
\vspace{-3pt}
\begin{equation}
\begin{aligned}
\mathcal{L}_{\mathrm{E}}(\boldsymbol{\hat{x}}_t)  &= f^{y}_T (\boldsymbol{\hat{x}}_t)\mathbf{E}(\boldsymbol{\hat{x}}_t,f_T) \\
                                                  &= -\alpha f^{y}_T(\boldsymbol{\hat{x}}_t) \log \sum_{i}^{K} e^{f^i_T(\boldsymbol{\hat{x}}_t)/\alpha},
\end{aligned}
\vspace{-3pt}
\end{equation}
where $y$ denotes the target class, and $\alpha$ is the temperature coefficient.
The additional target confidence score $f^{y}_T(\boldsymbol{\hat{x}}_t)$ is incorporated to preserve semantic consistency in the synthesized image.

Furthermore, we incorporate BN regularization as an additional auxiliary guidance to enforce distributional consistency.
Unlike inversion-based methods~\cite{yinDreamingDistillDataFree2020, fangContrastiveModelInversion2021, fang100xFasterDataFree2022, tranNAYERNoisyLayer2024} that rely on BN statistics to directly reconstruct training images, our approach utilizes this regularization exclusively for aligning feature distributions, while preserving image fidelity through the use of an off-the-shelf diffusion model.
The formalization of BN regularization guidance is as follows:
\vspace{-5pt}
\begin{equation}
    \mathcal{L}_{\mathrm{BN}}(\boldsymbol{\hat{x}}_t) = \sum_{l=1}^{L} \left( \lVert \mu_l(\boldsymbol{\hat{x}}_t) - \mu_l \rVert_2 + \lVert \sigma^2_l(\boldsymbol{\hat{x}}_t) - \sigma^2_l \rVert_2 \right).
\vspace{-5pt}
\end{equation}

In summary, to synthesize ID data with sufficient fidelity, the conditional generation for the latent image is defined as:
\vspace{-2pt}
\begin{equation}
\boldsymbol{z}_{t-1} = s(\boldsymbol{z}_t, t, \boldsymbol{\epsilon}_{\phi})- \rho_t \nabla_{\boldsymbol{z}_t} \mathcal{L}_{\mathrm{BN}}(\boldsymbol{\hat{x}}_t) - \gamma_t \nabla_{\boldsymbol{z}_t} \mathcal{L}_{\mathrm{E}}(\boldsymbol{\hat{x}}_t),
\vspace{-2pt}
\end{equation}
where $\rho_t$ and $\gamma_t$ are scaling factors.
In practice, we observed that applying EDA during the early denoising stages of DDPM leads to mode collapse and artifacts. This issue arises because the early stages are predominantly dominated by noise, with insufficient semantic information. To address this, we define the time interval guided by EDA as $[\tau_{start}, \tau_{end}]$ and empirically set $\tau_{start} \geq 0.6T$.
The detailed implementation is shown in Algorithm~\ref{alg:prism}.

\vspace{-5pt}
\subsection{Diversified Prompt Engineering} 
Diversified Prompt Engineering (DPE) is dedicated to solving the limited diversity of constraint views caused by the naive prompt.
DPE consists of three complementary components: semantic disambiguation, which resolves class name ambiguities by incorporating superclasses or textual definitions to clarify the intended semantics; content diversification, which encourages the generation of semantically distinct visual scenarios around the same concept; and style diversification, which enriches the representation through varied stylistic expressions.
For semantic disambiguation, we follow the study \cite{sariyildiz2023fake} to leverage WordNet~\cite{miller1995wordnet} to enrich class names with their superclasses or textual definitions, resolving ambiguities (e.g., “crane” as a bird vs. a machine). This semantic context steers the LLMs toward generating prompts consistent with the intended category.
For content diversification, we prompt LLMs to generate multiple scene descriptions conditioned on a single category, with explicit instructions such as “be creative and avoid repetition”. A higher sampling temperature is also employed to enhance variability. 
For style diversification, the LLMs are further guided to produce varied stylistic forms, e.g., “oil painting,” “cartoon”, or “photorealistic rendering”, which has been widely validated~\cite{xuan2023distilling}.
Concretely, for each class, we prompt the LLM to generate $N_c$ content descriptions and combine them with $N_s$ style variations, resulting in $N_c \times N_s$ diversified prompts per class. 

Compared to handcrafted prompt design, DPE is fully automated through LLM-based prompt engineering, significantly reducing human effort while enhancing sample diversity. 
\vspace{-5pt}
\section{Experiment}
\textbf{Models and Datasets}.
We validate our method in both homogeneous (ResNet34~\cite{heDeepResidualLearning2016} -> ResNet18~\cite{heDeepResidualLearning2016}) and heterogeneous (ResNet34~\cite{heDeepResidualLearning2016} -> MobileNetV2~\cite{sandlerMobileNetV2InvertedResiduals2019}) settings, across various large-scale ImageNet subsets, including ImageNette~\cite{howard2019smaller}, ImageWoof~\cite{howard2019smaller}, ImageNet-100~\cite{tian2020contrastive} (IN-100), and ImageNet-1k~\cite{russakovsky2015imagenet} (IN-1K).\vspace{3pt} \\
\textbf{Implementation Details}.
For fair comparison, we generate the same number of synthetic images for each method. 
For ImageNette, ImageWoof, and ImageNet100, we generate images of the same size as the training set, i.e., 10K, 10K, and 130K. 
For ImageNet, we follow the previous studies~\cite{yinDreamingDistillDataFree2020,fangContrastiveModelInversion2021,fang100xFasterDataFree2022,tranNAYERNoisyLayer2024} to generate 140K synthetic samples.
We use the SD V1.5 diffusion model\footnote{https://huggingface.co/stable-diffusion-v1-5/stable-diffusion-v1-5} to generate synthetic images and DeepSeek-V3.1~\footnote{https://huggingface.co/deepseek-ai/DeepSeek-V3.1} to achieve prompt diversification.
The experimental evaluations are conducted on 4x NVIDIA Hopper H100 80GiB GPU.

\begin{table}[h]
\centering
\caption{Results on large-scale ImageNet subsets.
\tred{Red} scores denote higher performance than the models trained on real images.}
\label{tab:main}
\renewcommand{\arraystretch}{0.95}
\resizebox{\columnwidth}{!}{%
\begin{tabular}{c|cc|cc|cc|c}
\toprule
Dataset     & \multicolumn{2}{c|}{ImageNette} & \multicolumn{2}{c|}{ImageWoof} & \multicolumn{2}{c|}{IN-100} & \multicolumn{1}{c}{IN-1K} \\ \midrule
Student & RN18  & MBNV2  & RN18  & MBNV2  & RN18  & MBNV2  & RN18 \\
  Tea.     & 94.2      & 94.2         & 89.8      & 89.8         & 85.3      & 85.3         & 73.3     \\
  Stu.      & 92.5      & 93.1         & 88.0      & 88.2         & 83.4      & 84.0         & 69.7     \\ \midrule
\multicolumn{8}{l}{\textit{Inversion-based}}                                                        \\ \midrule
CMI~\cite{fangContrastiveModelInversion2021}     & 91.7      & 90.3         & 86.3      & 86.8         & 68.4      & 65.9         & 37.2     \\
Fast50~\cite{fang100xFasterDataFree2022}  & 93.3      & 91.4         & \underline{88.5}      & \underline{87.8}         & 74.8      & 73.9         & 39.5     \\
NAYER~\cite{tranNAYERNoisyLayer2024}    & 92.8      & 91.0         & 87.7      & 87.5         & 72.1      & 69.7         & 38.6     \\
\midrule
\multicolumn{8}{l}{\textit{Diffusion-based}}                                                        \\ \midrule
DM-KD~\cite{liSyntheticDataDiffusion2023}   & 93.7      & 91.4         & 87.9      & 87.1         & 77.6      & 75.4         & 53.6     \\
GenQ~\cite{liGenQQuantizationLow2024}& \underline{94.2}      & \underline{92.8}         & 88.4      & 87.4         & \underline{80.1}      & \underline{79.7}         & \underline{55.9}     \\
 \textbf{PRISM} & \tred{\textbf{94.7}}  & \tred{\textbf{93.2}} & \tred{\textbf{89.5}} & \textbf{88.2} & \tred{\textbf{83.5}}  & \textbf{82.3}  & \textbf{65.1}                  \\ 
 $\Delta$  & \tgreen{0.5 $\uparrow$} &  \tgreen{0.4 $\uparrow$} & \tgreen{1.0 $\uparrow$} & \tgreen{0.4 $\uparrow$} & \tgreen{2.4 $\uparrow$} & \tgreen{2.6 $\uparrow$} & \tgreen{9.2 $\uparrow$} \\               \bottomrule
\end{tabular}
}
\vspace{-5pt}
\end{table}
\begin{table}[t]
\centering
\caption{Evaluation of domain generalization.
IN-100-S and IN-1K-S indicate DFKD settings.
*IN-R on IN-100 is calculated by the common classes of the two datasets.
\tred{Red} scores denote higher performance than the models trained on real images.}
\label{tab:generalize}
\renewcommand{\arraystretch}{0.90}
\resizebox{\columnwidth}{!}{%
\begin{tabular}{llcccccc}
\toprule
\multirow{2}{*}{Dataset} & \multirow{2}{*}{Method} & \multicolumn{2}{c}{IN-V2} & \multicolumn{2}{c}{IN-R*} & \multicolumn{2}{c}{IN-Sketch} \\
\cmidrule(lr){3-4} \cmidrule(lr){5-6} \cmidrule(lr){7-8}
 & & {Top-1} & {Top-5} & {Top-1} & {Top-5} & {Top-1} & {Top-5} \\
\midrule
 \multirow{1}{*}{IN-100}
 & RN18 & 70.7 & 88.3 & 35.6 & 53.8 & 30.0 & 50.3 \\
 \midrule
 \multirow{3}{*}{IN-100-S}
 & NAYER~\cite{tranNAYERNoisyLayer2024} & 59.6 & 83.4 & 27.1 & 45.6 & 18.6 & 38.6 \\
 & GenQ~\cite{liGenQQuantizationLow2024} & \underline{69.2} & \underline{87.0} & \underline{34.4} & \underline{53.2} & \underline{29.6} & \underline{49.0} \\
 & \textbf{PRISM} & \tred{\textbf{72.1}} & \tred{\textbf{90.3}} & \tred{\textbf{40.5}} & \tred{\textbf{58.8}} & \tred{\textbf{35.2}} & \tred{\textbf{54.3}} \\
\bottomrule
\toprule
\multirow{1}{*}{IN-1K}
 & RN18 & 57.3 & 79.9 & 20.4 & 34.5 & 20.2 & 37.2 \\
 \midrule
\multirow{3}{*}{IN-1K-S}
 & NAYER~\cite{tranNAYERNoisyLayer2024} & 35.4 & 45.2 & 10.0 & 21.4 & 8.8 & 21.5 \\
 & GenQ~\cite{liGenQQuantizationLow2024} & \underline{44.7} & \underline{70.3} & \underline{16.4} & \underline{30.0} & \underline{12.5} & \underline{31.4} \\
 & \textbf{PRISM} & \textbf{53.5} & \textbf{78.2} & \tred{\textbf{21.2}} & \tred{\textbf{34.8}} & \textbf{19.4} & \textbf{36.8} \\
\bottomrule
\end{tabular}
}
\vspace{-13pt}
\end{table}
\subsection{Main Results}
Table~\ref{tab:main} shows the performance comparison of our method with Inversion-based and Diffusion-based methods, across both homogeneous and heterogeneous distillation settings.
Since ImageNette and ImageWoof contain only ten classes, resulting in relatively compact feature distributions, inversion-based methods perform comparably to diffusion-based approaches.
In contrast, on more complex datasets such as IN-100 and IN-1K, diffusion-based methods exhibit a clear advantage, which can be attributed to the strong prior provided by the diffusion model.
Nevertheless, our PRISM achieves state-of-the-art performance across all datasets and even outperforms students on real training data in several settings.
It is noteworthy that our approach significantly outperforms GenQ on IN-100 (2.5$\uparrow$) and IN-1K (9.2$\uparrow$), demonstrating the capability to generalize in real-world scenarios.
\vspace{-8pt}
\subsection{Domain Generalization}
\label{sec:dg}
We also evaluate the performance of PRISM on challenging datasets: ImageNet-V2~\cite{recht2019imagenet} (IN-V2), ImageNet-Sketch~\cite{wang2019learning} (IN-Sketch), and ImageNet-R~\cite{hendrycks2021many} (IN-R), which contain OOD images, designed to assess model resilience to domain shifts and adversarial examples. 

Results in Table~\ref{tab:generalize} show that: on IN-100, students trained with our method consistently outperform previous approaches and even surpass the student trained on the IN-100 training data. 
This improvement stems not only from the strong prior of the diffusion model but also from our method’s ability to discover a richer diversity of visual patterns beyond the original data, significantly enhancing domain generalization. 
On the more challenging IN-1K dataset, our method clearly outperforms the second-best, GenQ. Although slightly below a ResNet-18 trained on the full IN-1K (1.28M images), our model uses only ~140k synthetic images, about one-tenth of the data. 
Remarkably, PRISM surpasses the IN-1K-trained student on the IN-R benchmark, demonstrating its strong generalization capability.
\begin{table}[t]
\caption{Ablation studies on IN-100.
Settings: RN34->RN18.}
\centering
\label{tab:ablation}
\footnotesize
\renewcommand{\arraystretch}{0.5}
\setlength{\tabcolsep}{1.2mm}
\begin{tabular}{@{}cc|c|cccccc|c@{}}
\toprule
\multicolumn{2}{c|}{Settings} &
  Baseline &
  \multicolumn{3}{c}{w/ DPE} &
  \multicolumn{3}{c|}{w/ EDA} &
  PRISM \\ \midrule
\multicolumn{1}{c|}{\multirow{2}{*}{EDA}} & $\mathcal{L}_{\mathrm{BN}}$ & \XSolidBrush & \XSolidBrush & \Checkmark & \XSolidBrush     & \Checkmark & \Checkmark & \Checkmark & \Checkmark \\
\multicolumn{1}{c|}{}    & $\mathcal{L}_{\mathrm{E}}$  & \XSolidBrush & \XSolidBrush & \XSolidBrush     & \Checkmark & \Checkmark & \Checkmark & \Checkmark & \Checkmark \\ \midrule
\multicolumn{1}{c|}{\multirow{2}{*}{DPE}} &
  C.D. &
  \XSolidBrush &
  \Checkmark &
  \Checkmark &
  \Checkmark &
  \XSolidBrush &
  \Checkmark &
  \XSolidBrush &
  \Checkmark \\
\multicolumn{1}{c|}{} &
  S.D. &
  \XSolidBrush &
  \Checkmark &
  \Checkmark &
  \Checkmark &
  \XSolidBrush &
  \XSolidBrush &
  \Checkmark &
  \Checkmark \\ \midrule
  \multicolumn{2}{c|}{ACC} &
  77.9 &
  80.3 &
  81.7 &
  82.9 &
  80.8 &
  83.2 &
  83.1 &
  \textbf{83.5} \\ \bottomrule
\end{tabular}%
\vspace{-5pt}
\end{table}
\begin{table}[t]
\centering
\caption{Quantitative comparison of synthetic data distributions against the IN-100 training set.}
\label{tab:fid}
\footnotesize
\renewcommand{\arraystretch}{1.05}
\setlength{\tabcolsep}{2.5mm}
\begin{tabular}{lcccc}
\toprule
Method & FID $\downarrow$  & Precision $\uparrow$ & Recall $\uparrow$  & Acc $\uparrow$\\
\midrule
NAYER~\cite{tranNAYERNoisyLayer2024}  & 24.4 & \textbf{96.7}      & 51.2 & 72.1  \\
GenQ~\cite{liGenQQuantizationLow2024}   & \underline{11.2} & 73.5     & \underline{60.8} & \underline{80.1}  \\
\textbf{PRISM}  & \textbf{5.9}  & \underline{81.6}      & \textbf{66.0}  & \textbf{83.5} \\ \bottomrule  
\end{tabular}%
\vspace{-6pt}
\end{table}
\begin{figure}[!t]
    \centering
    \includegraphics[width=\linewidth]{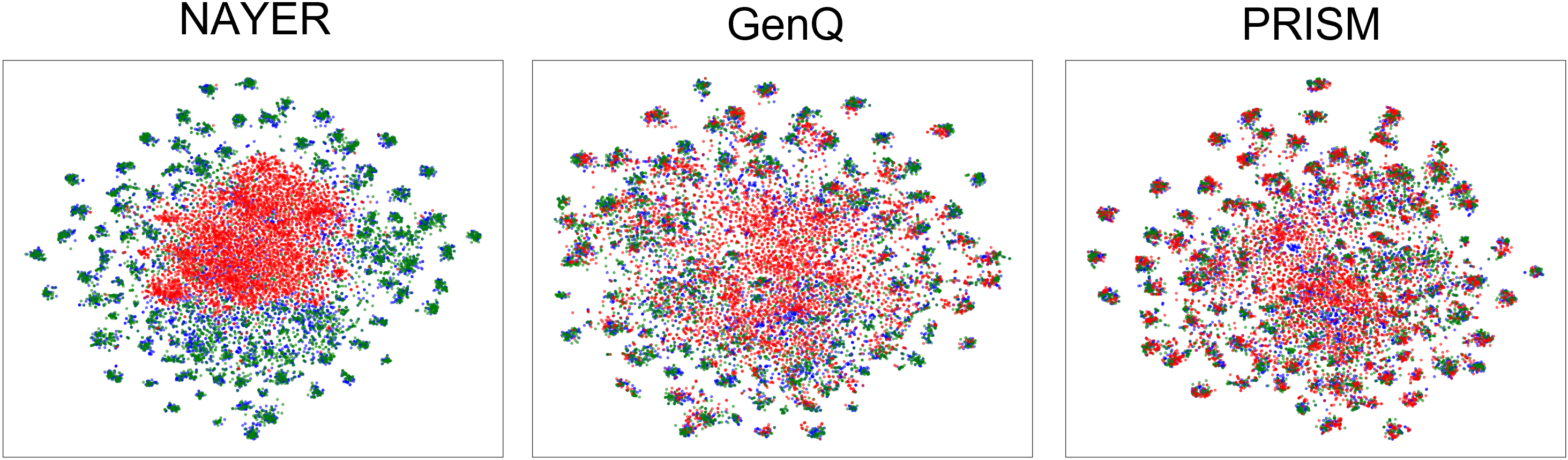}
    \caption{Coverage of synthetic distribution on IN-100.
    The t-SNE visualization of the teacher's features on the training (\tblue{Blue}), validation (\tgreen{Green}), and synthetic (\tred{Red}) datasets, respectively.
    }
    \label{fig:tsne}
    \vspace{-10pt}
\end{figure}
\vspace{-5pt}
\subsection{Method Analysis}
\noindent\textbf{Ablation Studies}.
We conduct ablation experiments on IN-100 to evaluate the individual contributions of EDA and DPE, as summarized in Table~\ref{tab:ablation}.
The results show that both components enhance student performance, indicating that improving either precision or recall alone offers substantial benefits. 
By integrating both mechanisms, our PRISM achieves a significant overall performance improvement.\vspace{3pt}

\noindent\textbf{Synthesis Quality}.
The qualitative results on the synthetic data are shown in Fig.~\ref{fig:visualize}, demonstrating that PRISM exhibits both semantic consistency and diversity.
Table~\ref{tab:fid} quantitatively compares synthetic distributions using FID, where our method achieves the closest match to the real distribution. 
While NAYER excels in precision, it produces clustered synthetic samples with limited recall. 
GenQ improves diversity through filtering but suffers from significant distribution shifts, reducing precision. 
In contrast, our PRISM strikes a superior precision-recall balance, ensuring both distribution consistency and diversity. 
This is further supported by Fig.~\ref{fig:tsne}, where NAYER’s feature distribution (red) under-covers the true training and validation samples, GenQ expands coverage with some distribution shifts, and our method effectively covers all clusters of the true distribution.

\vspace{-5pt}
\section{Conclusion}
In this paper, we propose PRISM, a precision–recall informed synthesis method for data-free knowledge distillation, addressing the critical challenges in synthesizing high-fidelity and diverse training samples for large-scale image datasets.
PRISM integrates Energy-guided Distribution Alignment to enforce strict alignment to the in-distribution manifold, thus enhancing the precision of the synthesized data, and Diversified Prompt Engineering driven by large language models to enrich sample diversity and improve recall.
Extensive experiments on ImageNet subsets demonstrate that PRISM significantly outperforms existing DFKD methods in both distillation and generalization.

\section{Acknowledgements}
This work was supported by the Sichuan Science and Technology Program under Grant Number 2024ZDZX0011.
\bibliographystyle{IEEEbib}
\bibliography{refs}


\end{document}